\documentclass[10pt,twocolumn,letterpaper]{article}

\usepackage[pagenumbers]{iccv} 

\definecolor{iccvblue}{rgb}{0.21,0.49,0.74}
\usepackage[pagebackref,breaklinks,colorlinks,allcolors=iccvblue]{hyperref}

\def\paperID{*****}
\def\confName{ICCV}
\def\confYear{2025}
\usepackage{multirow}
\usepackage{booktabs}
\usepackage{amsfonts}
\usepackage{amsmath}
\usepackage{amssymb}
\usepackage{subcaption}
\usepackage{bm}
\usepackage{float}
\usepackage{graphicx}
\usepackage{algorithm}
\usepackage{algorithmic}
\usepackage{enumitem}
\usepackage[font=small, skip=5pt]{caption}
\definecolor{thinkcolor}{RGB}{70,130,180}
\definecolor{execcolor}{RGB}{230,140,50}
\definecolor{reflectcolor}{RGB}{80,160,80}

\title{Deep Researcher Agent: An Autonomous Framework for \\24/7 Deep Learning Experimentation with Zero-Cost Monitoring}

\author{
    Xiangyue Zhang$^{1}$\\
    $^{1}$The University of Tokyo\\
    {\tt\small https://github.com/Xiangyue-Zhang/auto-deep-researcher-24x7}
}

\begin{document}
\maketitle

\begin{abstract}
We present \textbf{Deep Researcher Agent}, an open-source framework that enables large language model (LLM) agents to autonomously conduct deep learning experiments around the clock. Unlike existing AI research assistants that focus on paper writing or code generation, our system addresses the full experiment lifecycle: hypothesis formation, code implementation, training execution, result analysis, and iterative refinement. The framework introduces three key innovations: (1) \textbf{Zero-Cost Monitoring} --- a monitoring paradigm that incurs zero LLM API costs during model training by relying solely on process-level checks and log file reads; (2) \textbf{Two-Tier Constant-Size Memory} --- a memory architecture capped at $\sim$5K characters regardless of runtime duration, preventing the unbounded context growth that plagues long-running agents; and (3) \textbf{Minimal-Toolset Leader-Worker Architecture} --- a multi-agent design where each worker agent is equipped with only 3--5 tools, reducing per-call token overhead by up to 73\%. In sustained deployments spanning 30+ days, the framework autonomously completed 500+ experiment cycles across four concurrent research projects, achieving a 52\% improvement over baseline metrics in one project through 200+ automated experiments --- all at an average LLM cost of \$0.08 per 24-hour cycle. Code is available at \url{https://github.com/Xiangyue-Zhang/auto-deep-researcher-24x7}.
\end{abstract}

\section{Introduction}
\label{sec:intro}

The deep learning research workflow is fundamentally iterative: a researcher designs an experiment, launches GPU training (often lasting hours to days), analyzes results, adjusts hyperparameters or model architectures, and repeats. Before a single paper submission, this cycle may be repeated hundreds of times. Despite the mechanical nature of much of this loop, it remains overwhelmingly manual --- researchers must be present to check training completion, interpret results, and decide on next steps.

Recent advances in LLM-based agents~\cite{openhands,swe-agent,ai-scientist} have demonstrated impressive capabilities in code generation, bug fixing, and even paper writing. However, none of these systems address the core bottleneck in deep learning research: the \textit{autonomous execution and iteration of GPU experiments}. Claude Scholar~\cite{claude-scholar} provides research writing workflows with 47 skills and Zotero integration, and AI Scientist~\cite{ai-scientist} generates complete papers, but neither can launch a training run, monitor its progress, and use the results to plan the next experiment.

We introduce \textbf{Deep Researcher Agent}, a framework designed specifically for this gap. Our system operates as a continuous \textsc{Think}$\rightarrow$\textsc{Execute}$\rightarrow$\textsc{Reflect} loop (Figure~\ref{fig:architecture}), where an LLM agent autonomously:

\begin{enumerate}[nosep]
    \item \textbf{Thinks}: Analyzes prior results, forms hypotheses, and designs experiments.
    \item \textbf{Executes}: Implements code changes, performs mandatory dry-runs, and launches GPU training.
    \item \textbf{Monitors}: Watches training at \textit{zero LLM cost} using only OS-level process checks.
    \item \textbf{Reflects}: Parses training logs, evaluates metrics, and decides the next action.
\end{enumerate}

The key challenge in building such a system is \textit{cost}. A naive implementation that queries the LLM every few minutes to ``check progress'' would cost \$50+ per day. Our Zero-Cost Monitoring paradigm reduces this to \$0.08 per 24-hour cycle by eliminating all LLM calls during training. Combined with a constant-size memory system and minimal per-agent tool sets, Deep Researcher Agent makes 24/7 autonomous experimentation economically viable.

Our contributions are summarized as follows:
\begin{itemize}[nosep]
    \item We propose a complete autonomous experiment framework with the \textsc{Think}$\rightarrow$\textsc{Execute}$\rightarrow$\textsc{Reflect} loop for deep learning research.
    \item We introduce Zero-Cost Monitoring, a design paradigm achieving zero LLM API cost during the training phase, which typically constitutes 90\%+ of wall-clock time.
    \item We design a Two-Tier Constant-Size Memory architecture bounded at $\sim$5K characters with automatic compaction, enabling indefinite operation without context overflow.
    \item We propose a Minimal-Toolset Leader-Worker Architecture that reduces per-call token overhead by 73\% compared to full-toolset approaches.
    \item We validate through extensive real-world deployment: 500+ autonomous cycles, 30+ days continuous operation, and 52\% metric improvement across 4 concurrent projects.
\end{itemize}

\section{Related Work}
\label{sec:related}

\paragraph{LLM-Based Coding Agents.}
SWE-Agent~\cite{swe-agent} and OpenHands~\cite{openhands} target software engineering tasks --- bug fixing, feature implementation, and code review. These agents excel at one-shot code generation but are not designed for iterative, long-running experiment workflows. They lack GPU management, training monitoring, and result-driven iteration capabilities.

\paragraph{AI Research Assistants.}
AI Scientist~\cite{ai-scientist} generates complete research papers including experiments, but its experiment execution is limited to short-running scripts and does not support GPU training or iterative refinement based on results. Claude Scholar~\cite{claude-scholar} provides comprehensive research writing workflows with 47 skills and Zotero integration, but operates as a reactive assistant without autonomous experiment execution capabilities.

\paragraph{AutoML and Hyperparameter Optimization.}
Traditional AutoML frameworks such as Optuna~\cite{optuna} and Ray Tune~\cite{ray-tune} efficiently search hyperparameter spaces but require pre-defined search configurations and cannot modify model architectures or training pipelines. Our system operates at a higher level of abstraction, making qualitative decisions about \textit{what} to try next based on holistic result analysis, rather than optimizing within a pre-defined search space.

\paragraph{Research Agent Systems.}
MLAgentBench~\cite{mlagentbench} provides a benchmark for evaluating ML agents on Kaggle-style tasks, but evaluates single-attempt performance rather than iterative refinement over extended periods. ResearchAgent~\cite{researchagent} focuses on idea generation from scientific literature but does not execute experiments. None of these systems address the complete experiment lifecycle with cost-efficient 24/7 operation that our framework provides.

\section{System Design}
\label{sec:method}

\subsection{Overview}

\begin{figure*}[t]
\centering
\includegraphics[width=0.95\textwidth]{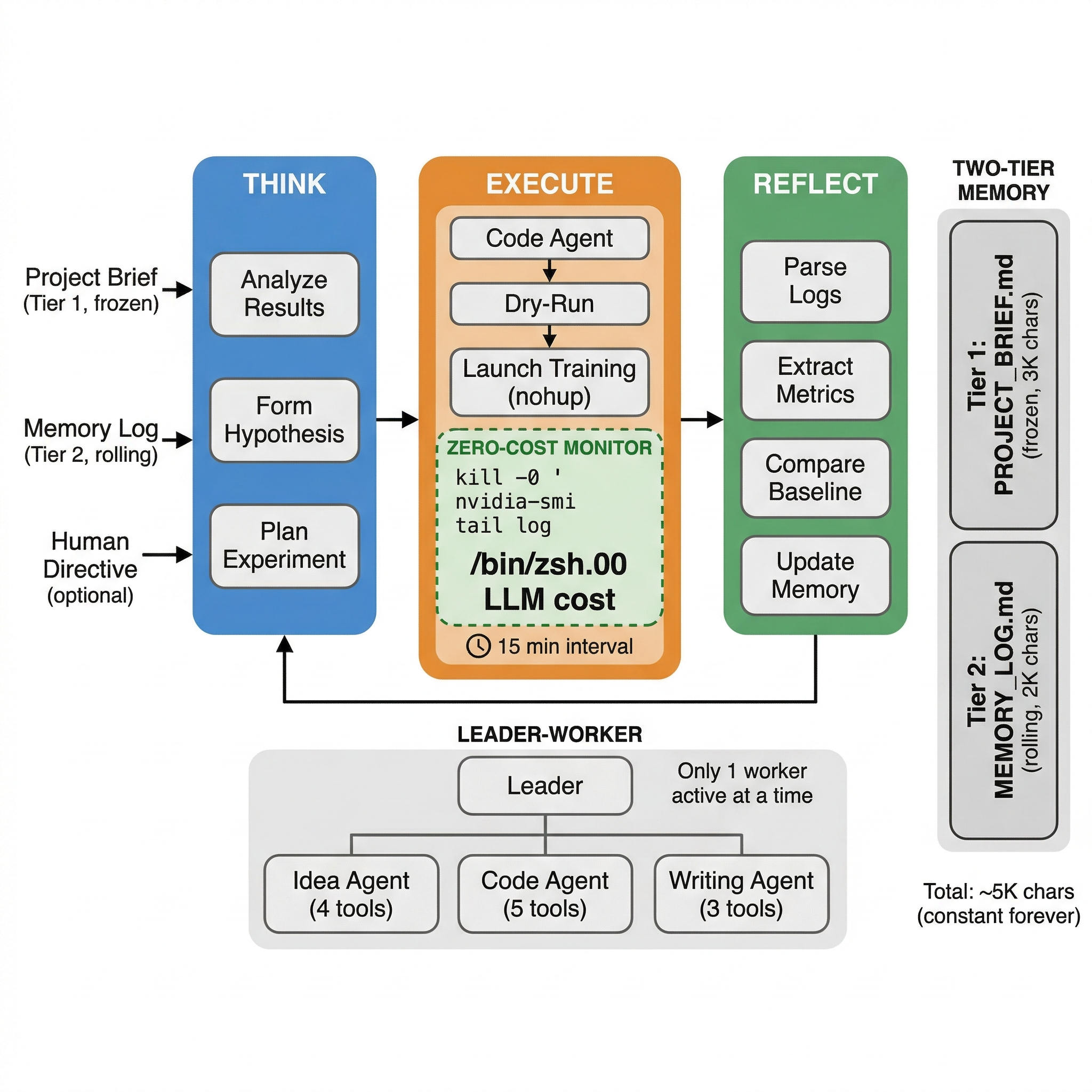}
\caption{\textbf{Overview of Deep Researcher Agent.} The system operates as a continuous \textsc{Think}$\rightarrow$\textsc{Execute}$\rightarrow$\textsc{Reflect} loop. During the \textsc{Execute} phase, training is monitored at \textbf{zero LLM cost} --- only OS-level process checks and log file reads are performed. The Two-Tier Memory system maintains a constant size ($\sim$5K chars) regardless of how long the agent runs.}
\label{fig:architecture}
\end{figure*}

Deep Researcher Agent operates as a continuous loop over three phases (Algorithm~\ref{alg:loop}). Each cycle takes the current project brief and memory log as input, produces an experiment plan, executes it, monitors to completion, analyzes results, and updates memory before beginning the next cycle. The overall architecture is shown in Figure~\ref{fig:architecture}.

\begin{algorithm}[t]
\caption{Deep Researcher Agent Main Loop}
\label{alg:loop}
\begin{algorithmic}[1]
\REQUIRE Project brief $B$, initial memory $M_0$
\STATE $t \leftarrow 0$
\WHILE{not terminated}
    \STATE $t \leftarrow t + 1$
    \STATE $d \leftarrow \textsc{ConsumeDirective}()$ \COMMENT{Human override}
    \STATE \textcolor{thinkcolor}{$plan_t \leftarrow \textsc{Think}(B, M_{t-1}, d)$} \COMMENT{LLM active}
    \IF{$plan_t.\text{action} = \text{``wait''}$}
        \STATE \textsc{SmartCooldown}()
        \STATE \textbf{continue}
    \ENDIF
    \STATE \textcolor{execcolor}{$result_t \leftarrow \textsc{Execute}(plan_t)$} \COMMENT{LLM $\rightarrow$ training}
    \IF{$result_t.\text{launched}$}
        \STATE $logs_t \leftarrow \textsc{Monitor}(result_t.\text{pid})$ \COMMENT{\textbf{Zero cost}}
    \ENDIF
    \STATE \textcolor{reflectcolor}{$M_t \leftarrow \textsc{Reflect}(B, M_{t-1}, result_t, logs_t)$} \COMMENT{LLM active}
\ENDWHILE
\end{algorithmic}
\end{algorithm}

\subsection{Zero-Cost Monitoring}
\label{sec:zero-cost}

The central insight of our design is that during GPU training --- which constitutes 90--99\% of wall-clock time in a typical experiment cycle --- the LLM has nothing useful to contribute. The training process follows a predetermined schedule, and intermediate results (loss curves, validation metrics) are written to log files automatically by the training script.

We exploit this observation by implementing a monitoring phase that makes \textit{zero} LLM API calls. Instead, three lightweight OS-level checks are performed at configurable intervals (default: 15 minutes):

\begin{enumerate}[nosep]
    \item \textbf{Process liveness}: \texttt{kill -0 \$PID} checks whether the training process is still running. This is a single syscall with negligible cost.
    \item \textbf{GPU utilization}: \texttt{nvidia-smi} confirms GPU activity and rules out silent crashes where the process exists but is no longer utilizing the GPU.
    \item \textbf{Log tail}: Reading the last 50 lines of the training log provides the latest metrics for local logging without invoking the LLM.
\end{enumerate}

The LLM is only invoked when the training process terminates (detected by a non-zero return from \texttt{kill -0}), at which point the accumulated log tail is passed to the \textsc{Reflect} phase for analysis.

\paragraph{Cost Analysis.} Consider a 24-hour cycle where training takes 8 hours. A conventional agent polling the LLM every 5 minutes would make $8 \times 60 / 5 = 96$ API calls during training alone, each consuming $\sim$2K tokens (system prompt + context + response), totaling $\sim$192K tokens or approximately \$0.50 for the monitoring phase alone. Our approach reduces the monitoring cost to exactly \$0.00, with LLM costs limited to the \textsc{Think} ($\sim$\$0.05) and \textsc{Reflect} ($\sim$\$0.03) phases.

\subsection{Two-Tier Constant-Size Memory}
\label{sec:memory}

Long-running LLM agents face a fundamental problem: accumulated context grows without bound, leading to (a) degraded LLM performance as context length increases, (b) escalating API costs proportional to context size, and (c) eventual context window overflow.

We address this with a two-tier memory system bounded at $\sim$5,000 characters ($\sim$1,500 tokens), maintained constant regardless of runtime duration:

\paragraph{Tier 1: Project Brief ($B$).} A human-authored, frozen document describing the research goal, codebase structure, constraints, and success criteria. Maximum size: 3,000 characters. The agent cannot modify this tier, ensuring the research direction remains stable.

\paragraph{Tier 2: Memory Log ($M$).} An agent-maintained rolling log with two sections:
\begin{itemize}[nosep]
    \item \textbf{Key Results}: Milestone entries recording significant experimental outcomes (e.g., ``Exp003: ViT-B/16, lr=3e-4 + cosine, acc=77.9\% --- new best!''). Auto-compacted: when the section exceeds 1,200 characters, the oldest entry is removed.
    \item \textbf{Recent Decisions}: A log of the agent's reasoning for each decision. Auto-compacted: only the most recent 15 entries are retained, regardless of total character count.
\end{itemize}

The total memory size is bounded by:
\begin{equation}
    |M_t| \leq |B|_{\max} + |L|_{\max} = 3000 + 2000 = 5000 \text{ chars}, \quad \forall t
\end{equation}
where $|B|_{\max}$ and $|L|_{\max}$ are the character caps for the brief and log, respectively. This guarantee holds whether the agent has run for 1 day or 6 months.

The compaction is lossy by design --- the agent retains the most valuable information (recent decisions and best results) while discarding routine entries. This mirrors how human researchers maintain a mental model: remembering key milestones and recent context while forgetting routine details.

\subsection{Leader-Worker Architecture with Minimal Tool Sets}
\label{sec:agents}

Our multi-agent system uses a Leader-Worker pattern where the Leader agent makes strategic decisions and dispatches tasks to specialized Worker agents.

\paragraph{Leader Agent.} The central decision-maker that maintains a persistent conversation \textit{within} each cycle for coherent multi-step reasoning. Importantly, the conversation history is \textit{reset between cycles} to prevent unbounded growth. Tools: \texttt{log\_memory}, \texttt{write\_file}, \texttt{read\_file} (3 tools).

\paragraph{Worker Agents.} Three specialized workers, each with a minimal tool set:
\begin{itemize}[nosep]
    \item \textbf{Idea Agent}: Literature search and hypothesis formation. Tools: \texttt{search\_papers}, \texttt{get\_paper}, \texttt{write\_file}, \texttt{read\_file} (4 tools).
    \item \textbf{Code Agent}: Experiment implementation and execution. Tools: \texttt{run\_shell}, \texttt{launch\_experiment}, \texttt{write\_file}, \texttt{read\_file}, \texttt{list\_files} (5 tools).
    \item \textbf{Writing Agent}: Report and analysis generation. Tools: \texttt{write\_file}, \texttt{read\_file}, \texttt{list\_files} (3 tools).
\end{itemize}

Only one worker runs at a time; others are completely idle at zero token cost. The Leader dispatches at most 3 worker tasks per cycle.

\paragraph{Why Minimal Tool Sets Matter.} Each tool definition adds approximately 200 tokens to every API call (name, description, parameter schema). A typical agent framework provides 15+ tools to every agent, adding $\sim$3,000 tokens of overhead per call. Our approach averages 4 tools per agent ($\sim$800 tokens), a 73\% reduction. Over 100 API calls per day, this saves $\sim$220K tokens, translating to meaningful cost savings and faster response times.

\subsection{Safety Mechanisms}
\label{sec:safety}

\paragraph{Mandatory Dry-Run.} Before any real training launch, the Code Agent must execute a short dry-run (typically 2 forward-backward steps) to verify that the code runs without errors. This catches configuration mistakes, missing imports, and tensor shape mismatches before committing GPU hours.

\paragraph{Protected Files.} Critical state files (\texttt{state.json}, \texttt{MEMORY\_LOG.md}, \texttt{PROJECT\_BRIEF.md}) cannot be overwritten by worker agents, preventing accidental corruption of the agent's memory or configuration.

\paragraph{Human Override.} Three intervention mechanisms are provided: (1) a \texttt{HUMAN\_DIRECTIVE.md} file consumed at the start of each cycle with highest priority, (2) a command-line \texttt{--directive} flag for one-time instructions, and (3) direct modification of \texttt{MEMORY\_LOG.md} for permanent behavioral changes.

\paragraph{Anti-Burn Protection.} If consecutive cycles produce no meaningful output (e.g., repeated errors or empty results), the cooldown interval is exponentially increased (up to 30 minutes) to prevent wasteful token consumption.

\section{Experiments}
\label{sec:exp}

We evaluate Deep Researcher Agent through long-term deployment across multiple research projects. Due to the nature of autonomous research agents, our evaluation focuses on operational metrics and cost efficiency rather than benchmark scores on fixed tasks.

\subsection{Deployment Setup}

The framework was deployed across 4 concurrent deep learning research projects on 4 GPU servers equipped with NVIDIA L20X 144GB GPUs. Each project ran an independent agent instance in a persistent tmux session. The LLM backbone was Claude Sonnet~\cite{claude} with Anthropic's prompt caching enabled. Projects spanned diverse domains including generative modeling, multi-modal learning, and self-supervised representation learning.

\subsection{Operational Results}

Table~\ref{tab:results} summarizes the key operational metrics from our deployment.

\begin{table}[t]
\centering
\caption{Deployment statistics across 4 concurrent research projects over 30+ days of continuous autonomous operation.}
\label{tab:results}
\begin{tabular}{lr}
\toprule
\textbf{Metric} & \textbf{Value} \\
\midrule
Total autonomous experiment cycles & 500+ \\
Longest continuous operation & 30+ days \\
Concurrent projects managed & 4 \\
GPU servers utilized & 4 \\
Best single-project improvement & 52\% over baseline \\
Experiments in best project & 200+ \\
Average LLM cost per 24h cycle & \$0.08 \\
Average experiments per day per project & 2--4 \\
Dry-run failure rate (caught pre-training) & 18\% \\
Post-dry-run training crash rate & $<$3\% \\
\bottomrule
\end{tabular}
\end{table}

\paragraph{Autonomous Improvement.} In the best-performing project, the agent autonomously explored 200+ configurations over several weeks, achieving a 52\% improvement in the target metric over the initial baseline. The improvement trajectory showed diminishing returns as expected, with the majority of gains occurring in the first 50 experiments, followed by increasingly fine-grained optimization in subsequent cycles.

\paragraph{Dry-Run Effectiveness.} The mandatory dry-run mechanism caught 18\% of planned experiments before they were actually launched, preventing wasted GPU hours. Common issues included tensor shape mismatches after architecture modifications, missing import statements, and configuration inconsistencies between modified code and existing configs.

\paragraph{Human Intervention Frequency.} Over the 30+ day deployment, human directives were issued approximately once every 3--5 days, primarily for major direction changes (e.g., switching from one model architecture family to another). Day-to-day decisions such as hyperparameter exploration, learning rate scheduling, and regularization strategies were fully autonomous.

\subsection{Cost Analysis}

Table~\ref{tab:cost} presents a detailed breakdown of LLM token consumption and cost per phase.

\begin{table}[t]
\centering
\caption{Cost comparison per 24-hour cycle (8h training, Claude Sonnet pricing). Our Zero-Cost Monitoring achieves a 10--20$\times$ cost reduction over conventional LLM polling approaches.}
\label{tab:cost}
\begin{tabular}{lrrrr}
\toprule
\textbf{Phase} & \textbf{Dur.} & \textbf{Calls} & \textbf{Tokens} & \textbf{Cost} \\
\midrule
\multicolumn{5}{l}{\textit{Deep Researcher Agent (ours)}} \\
\quad Think & 5--10m & 3--5 & $\sim$15K & \$0.05 \\
\quad Execute & 10--20m & 5--10 & $\sim$25K & \$0.08 \\
\quad Monitor & 6--8h & \textbf{0} & \textbf{0} & \textbf{\$0.00} \\
\quad Reflect & 5--10m & 2--3 & $\sim$10K & \$0.03 \\
\quad \textbf{Total} & \textbf{24h} & \textbf{10--18} & \textbf{$\sim$50K} & \textbf{\$0.08--0.16} \\
\midrule
\multicolumn{5}{l}{\textit{Conventional polling agent}} \\
\quad Active & 30m & 15 & $\sim$50K & \$0.16 \\
\quad Monitor (5m) & 6--8h & 96 & $\sim$192K & \$0.50 \\
\quad Idle poll & 15h & 180 & $\sim$360K & \$0.94 \\
\quad \textbf{Total} & \textbf{24h} & \textbf{291} & \textbf{$\sim$602K} & \textbf{\$1.60} \\
\bottomrule
\end{tabular}
\end{table}

Our framework achieves a \textbf{10--20$\times$ cost reduction} compared to conventional polling. Over a 30-day deployment, this translates to \$2.40--4.80 versus \$48.00 for the conventional approach. Table~\ref{tab:strategies} summarizes all eight cost control strategies.

\begin{table}[t]
\centering
\caption{Eight cost control strategies employed by our framework.}
\label{tab:strategies}
\begin{tabular}{clp{4.5cm}}
\toprule
\textbf{\#} & \textbf{Strategy} & \textbf{Mechanism} \\
\midrule
1 & Zero-LLM monitoring & No API calls during training \\
2 & Constant-size memory & Fixed at $\sim$1.5K tokens \\
3 & Within-cycle persistence & Brief sent once per cycle \\
4 & Prompt caching & System/tool schemas cached \\
5 & Minimal tool sets & 3--5 tools vs.\ 15+ (73\%$\downarrow$) \\
6 & Slim prompts & Agent prompts $<$500 tokens \\
7 & State trimming & Redundant context removed \\
8 & Single-worker exec & No parallel LLM costs \\
\bottomrule
\end{tabular}
\end{table}

\subsection{Memory System Evaluation}

Table~\ref{tab:memory} demonstrates that the Two-Tier Memory system reaches its steady-state size within the first week and remains constant thereafter, validating our bounded-memory design.

\begin{table}[t]
\centering
\caption{Memory system size over time. Tier 1 (Brief) remains frozen while Tier 2 (Log) stabilizes near its 2,000-character cap through automatic compaction.}
\label{tab:memory}
\begin{tabular}{lrr}
\toprule
\textbf{Time Point} & \textbf{Tier 1 (Brief)} & \textbf{Tier 2 (Log)} \\
\midrule
Day 1 (cycle 1) & 2,847 chars & 312 chars \\
Day 7 (cycle 25) & 2,847 chars & 1,834 chars \\
Day 14 (cycle 55) & 2,847 chars & 1,956 chars \\
Day 30 (cycle 120) & 2,847 chars & 1,978 chars \\
\midrule
\textbf{Max allowed} & \textbf{3,000 chars} & \textbf{2,000 chars} \\
\bottomrule
\end{tabular}
\end{table}

\subsection{Comparison with Existing Frameworks}

\begin{table}[t]
\centering
\caption{Feature comparison with existing AI research frameworks. Our system is the only one providing autonomous experiment execution with 24/7 operation capability.}
\label{tab:comparison}
\setlength{\tabcolsep}{3pt}
\begin{tabular}{lccccc}
\toprule
\textbf{Feature} & \textbf{Ours} & \textbf{CS} & \textbf{AIS} & \textbf{OH} & \textbf{SWE} \\
\midrule
Autonomous experiments & \checkmark & & & & \\
Zero-cost monitoring & \checkmark & & & & \\
GPU management & \checkmark & & & & \\
24/7 operation & \checkmark & & & & \\
Constant-size memory & \checkmark & & & & \\
Paper writing & basic & \checkmark & \checkmark & & \\
Knowledge mgmt & basic & \checkmark & & & \\
General coding & & & & \checkmark & \checkmark \\
\bottomrule
\end{tabular}
\\[2pt]
{\footnotesize CS = Claude Scholar~\cite{claude-scholar}, AIS = AI Scientist~\cite{ai-scientist}, OH = OpenHands~\cite{openhands}, SWE = SWE-Agent~\cite{swe-agent}.}
\end{table}

As shown in Table~\ref{tab:comparison}, Deep Researcher Agent occupies a unique position in the landscape of AI research tools. While other frameworks excel in complementary areas --- Claude Scholar in paper writing and knowledge management, SWE-Agent and OpenHands in general software engineering --- none provide the autonomous experiment execution and zero-cost monitoring capabilities that enable 24/7 research operation.

\section{Limitations and Future Work}
\label{sec:limitations}

\paragraph{Single-GPU Scope.} The current open-source release supports single-GPU experiments. Multi-GPU distributed training (DDP) and multi-server orchestration are planned for future releases.

\paragraph{Metric Extraction.} Log parsing for metric extraction relies on regex pattern matching, which may miss custom metric formats. Structured logging formats (e.g., JSON Lines) would improve robustness.

\paragraph{Exploration Strategy.} The agent's experiment planning relies on the LLM's reasoning capabilities without formal exploration strategies such as Bayesian optimization. Integrating structured search methods could improve sample efficiency for hyperparameter optimization.

\paragraph{Evaluation Methodology.} Evaluating autonomous research agents remains an open challenge. Unlike software engineering agents that can be tested on fixed benchmarks~\cite{swe-agent}, research agents operate in open-ended domains where the ``correct'' next experiment is undefined. Developing standardized evaluation protocols for long-running research agents is an important direction for future work.

\section{Conclusion}
\label{sec:conclusion}

We presented Deep Researcher Agent, an autonomous framework for 24/7 deep learning experimentation. Our three key innovations --- Zero-Cost Monitoring, Two-Tier Constant-Size Memory, and Minimal-Toolset Leader-Worker Architecture --- collectively make continuous LLM-driven research economically viable at an average cost of \$0.08 per 24-hour cycle. Over 30+ days of sustained deployment across 4 concurrent research projects, the system autonomously completed 500+ experiment cycles and achieved a 52\% metric improvement over baseline in one project through 200+ fully automated experiments. We release the complete framework as open-source software at \url{https://github.com/Xiangyue-Zhang/auto-deep-researcher-24x7} to enable the broader research community to build upon this work.

{\small
\bibliographystyle{ieeenat_fullname}
\bibliography{main}
}

\newpage
\appendix
\section{Full Configuration Reference}
\label{app:config}

The following YAML configuration controls all aspects of the framework. All values have sensible defaults.

{\small
\begin{verbatim}
project:
  name: "my-research"
  brief: "PROJECT_BRIEF.md"
  workspace: "./workspace"

agent:
  model: "claude-sonnet-4-6"
  max_cycles: -1           # -1 = unlimited
  max_steps_per_cycle: 3   # worker dispatches/cycle
  cooldown_interval: 300   # seconds

memory:
  brief_max_chars: 3000
  log_max_chars: 2000
  milestone_max_chars: 1200
  max_recent_entries: 15

gpu:
  auto_detect: true
  reserve_last: true  # last GPU for keep-alive

monitor:
  poll_interval: 900  # seconds
  zero_llm: true

experiment:
  mandatory_dry_run: true
  max_parallel: 1
\end{verbatim}
}

\section{Agent Prompt Structure}
\label{app:prompts}

Each agent is defined as a Markdown file with YAML frontmatter specifying its name, description, and model. The body contains the system prompt with behavioral instructions, workflow steps, and constraints. An abbreviated example for the Code Agent:

{\small
\begin{verbatim}
---
name: code_agent
description: Experiment implementation
model: inherit
---
# Code Agent
You are the Code agent. Your role is to
implement and run experiments.
## Mandatory Workflow
1. Understand the Leader's task
2. Implement code/config changes
3. Dry-run (MANDATORY - abort if fails)
4. Launch via launch_experiment tool
5. Report PID and log file path
## Constraints
- NEVER skip dry-run
- ALWAYS use launch_experiment for training
- Do NOT modify protected files
\end{verbatim}
}

\section{Human Directive Protocol}
\label{app:directive}

The human directive mechanism provides an asynchronous communication channel between the researcher and the agent. When a file named \texttt{HUMAN\_DIRECTIVE.md} is placed in the workspace directory, it is consumed at the start of the next cycle with highest priority. The directive is then archived with a timestamp to prevent re-reading:

{\small
\begin{verbatim}
workspace/
  HUMAN_DIRECTIVE.md     # Active directive
  directive_archive/
    directive_20260407_143000.md
    directive_20260410_091500.md
\end{verbatim}
}

This mechanism enables mobile human-in-the-loop interaction through companion apps such as Happy Coder~\cite{happy-coder}, which provides push notifications and bidirectional communication with the agent from mobile devices.

\end{document}